# Optimizing TD3 for 7-DOF Robotic Arm Grasping: Overcoming Suboptimality with Exploration-Enhanced Contrastive Learning

Wen-Han Hsieh
*Department of Power Mechanical Engineering*
National Tsing Hua University
Hsinchu, Taiwan
hense54321@gmail.com

Jen-Yuan Chang
*Department of Power Mechanical Engineering*
National Tsing Hua University
Hsinchu, Taiwan
jychang@pme.nthu.edu.tw

*Abstract*—In actor-critic-based reinforcement learning algorithms such as Twin Delayed Deep Deterministic policy gradient (TD3), insufficient exploration of the spatial space can result in suboptimal policies when controlling 7-DOF robotic arms. To address this issue, we propose a novel Exploration-Enhanced Contrastive Learning (EECL) module that improves exploration by providing additional rewards for encountering novel states. Our module stores previously explored states in a buffer and identifies new states by comparing them with historical data using Euclidean distance within a K-dimensional tree (KDTree) framework. When the agent explores new states, exploration rewards are assigned. These rewards are then integrated into the TD3 algorithm, ensuring that the Q-learning process incorporates these signals, promoting more effective strategy optimization. We evaluate our method on the robosuite panda lift task, demonstrating that it significantly outperforms the baseline TD3 in terms of both efficiency and convergence speed in the tested environment.

## I. INTRODUCTION

TD3 is a robust model-free reinforcement learning (RL) algorithm that mitigates overestimation bias [1] through twin Q-networks and delayed policy updates [2]. Unlike other RL algorithms, such as A2C, A3C, DDPG, SAC, and PPO, TD3 provides a balanced trade-off between convergence speed and sample efficiency, making it ideal for controlling robotic arms. TD3's robustness and stability make it particularly effective in reducing training time and improving performance in dynamic environments.

Traditional control methods often require extensive and complex computations that are impractical in dynamic settings [3]. RL algorithms like TD3 offer an alternative by dynamically adapting to the environment and continuously optimizing strategies. However, TD3 still faces challenges in effectively exploring new strategies, resulting in suboptimal policies where the agent settles for locally optimal solutions and adaptability issues that prevent it from quickly adjusting to changes in dynamic environments [4].

Our approach introduces a contrastive learning module that distinguishes explored and unexplored states, providing additional rewards for discovering new regions. This targeted incentive accelerates exploration, improving TD3's adaptability and allowing robotic arms to better navigate dynamic tasks like precise object sorting or assembly. This module significantly enhances exploration efficiency and accelerates policy discovery, leading to quicker adaptation and reliable, high-precision control.

While previous intrinsic reward strategies often resulted in unstable policies or overgeneralization, our contrastive learning module directly balances exploration and exploitation through a structured reward system. Despite the potential challenges of tuning the reward scale, our solution greatly improves TD3's learning process, yielding superior policy optimization and control strategies for industrial robotic arms.

## II. RELATED WORK

### A. TD3 in Robotic arm grasping

TD3 is a popular algorithm for controlling 7-DOF robotic arms. TD3 addresses overestimation bias in Q-learning algorithms by employing twin Q-networks and delayed policy updates, improving stability and performance in continuous control tasks.

TD3's application in robotic arm grasping has shown promising results, enabling precise movements for grasping and lifting objects [5]. Its ability to handle high-dimensional action spaces and maintain sample efficiency makes it suitable for complex manipulation tasks. However, TD3 faces challenges in exploration, often resulting in suboptimal policies due to insufficient exploration.

Enhancements to TD3, such as integrating intrinsic motivation mechanisms and combining it with other RL algorithms, have been explored to encourage more diverse state exploration. These methods aim to overcome the exploration challenges and improve performance in robotic arm control tasks. Our proposed EECL module addresses this by providing additional rewards for novel state discovery, promoting more effective exploration and improved TD3 performance.

### B. Exploration Methods in RL

Exploration is a critical challenge in RL, where the agent must balance exploiting known information to maximize rewards and exploring new actions to discover better strategies. Count-based exploration methods [6] provide exploration bonuses for revisiting under-explored state-action pairs [7]. Leshem Choshen extended these ideas by constructing an additional MDP for exploration-value estimation [8].

Curiosity-driven methods, such as those by Deepak Pathak [9] and Yuri Burda [10], design intrinsic rewards to encourage exploration in continuous state tasks. These rewards are based on prediction errors or state novelty, promoting self-motivated exploration. Optimistic initialization techniques, such as those by Ian Osband [11], assign higher initial Q-values to rarely visited state-action pairs, naturally encouraging exploration.

In robotic arm control, effective exploration is crucial for discovering optimal policies. Our EECL module integrates these principles by providing exploration rewards within the TD3 framework, enhancing the agent's ability to explore novel states and improve performance.

*C. Contrastive Learning in DRL*

Contrastive learning is effective for unsupervised and self-supervised learning tasks, involving training models to distinguish between similar and dissimilar data points. In DRL, contrastive learning can enhance exploration by identifying and rewarding novel states.

Applications of contrastive learning in DRL, such as those by Aaron van den Oord [12] and Michael Laskin [13], have improved representation learning and sample efficiency. These methods train auxiliary networks to predict future states or rewards, using contrastive losses to encourage diverse state exploration.

Our EECL module builds on these ideas by incorporating contrastive learning into TD3. By storing explored states in a buffer and using a KDTree to identify novel states, the EECL module provides additional exploration rewards, accelerating exploration and leading to more effective strategy optimization in robotic arm control tasks.

### III. EXPLORATION-ENHANCED CONTRASTIVE LEARNING (EECL) MODULE

*A. Objective*

The goal of the Exploration-Enhanced Contrastive Learning (EECL) module is to enhance the exploration efficiency of the TD3 algorithm in controlling 7-DOF robotic arms. The module achieves this by providing additional rewards for encountering novel states, encouraging the agent to explore more comprehensively and avoid suboptimal policies.

Formally, let the state trajectory be denoted as $T = (s_1, s_2, \ldots, s_T)$, where each state $s_i$ is encountered during the learning process. The EECL module aims to maximize the exploration reward by identifying and rewarding novel states. To model the exploration process, we define the probability $p(s)$ of encountering a state $s$ as:

$$p(s) = \prod_{i=1}^{T} p(s_i | s_{i-1}, \ldots, s_1) \quad (1)$$

The exploration reward $r_e$ is computed for each novel state identified during the trajectory, and the total exploration reward is the sum of rewards for all novel states:

$$R_e = \sum_{t \in T} r_e(t) \quad (2)$$

The EECL module is trained to maximize this exploration reward, thereby improving the agent's ability to discover new strategies and optimize its policy effectively.

By promoting extensive exploration, the EECL module helps the agent to avoid local optima and enhances its adaptability to dynamic environments.

*B. State Storage and Management*

The EECL module employs a state buffer to store previously encountered states and a KDTree for efficient state management. This approach ensures that the module can quickly and accurately identify whether a new state is novel.

- **State Buffer**: Stores the states encountered during training. If the buffer reaches its maximum capacity, the oldest states are removed to accommodate new ones.
- **KDTree Construction**: Organizes the stored states into a KDTree data structure, which facilitates efficient nearest neighbor searches and comparisons.

The buffer and KDTree are updated as follows:

$$Buffer \leftarrow Buffer \cup \{s\} \quad (3)$$

$$KDTree \leftarrow KDTree(Buffer) \quad (4)$$

*C. Novel State Identification*

To identify novel states, the EECL module compares each new state with the states stored in the KDTree using Euclidean distance. A state is considered novel if its distance to the nearest neighbor in the KDTree exceeds a predefined threshold $\epsilon$.

Let $s$ be a new state and $s'$ be the nearest neighbor in the KDTree. The state $s$ is novel if:

$$||s - s'||_2 \geq \epsilon \quad (5)$$

If a state is determined to be novel, it is added to the state buffer and the KDTree is updated accordingly.

*D. Reward Assignment*

Once a novel state is identified, the EECL module assigns an exploration reward to the agent. This reward is initially set to a maximum value and decays over time, balancing exploration and exploitation.

The exploration reward can be formulated as:

$$r_e = r_{max} \times \gamma^n \quad (6)$$

where:

- $r_{max}$ is the initial maximum exploration reward.
- $\gamma$ is the reward decay factor.
- $n$ is the number of novel states discovered.

This decay mechanism ensures early exploration is heavily rewarded while the incentive gradually reduces as more of the state space is explored. By dynamically adjusting the reward based on state novelty, the EECL module guides the agent through a diverse state space, balancing exploration and exploitation to achieve optimal policy. This strategy improves the robustness and efficiency of the learned policy.

*E. Module Architecture*

The EECL module enhances TD3's exploration by storing states, identifying novel states, and assigning rewards. Using a KDTree for efficient nearest neighbor search, it quickly identifies novel states. The exploration reward mechanism dynamically adjusts based on state novelty, promoting balanced exploration and exploitation.

Algorithm provides a pseudocode overview of the integration of the EECL module with TD3.

**Algorithm**: TD3 with EECL:

Initialize critic networks $Q_1$, $Q_2$, and actor network $\pi$ with random parameters $\theta_1$, $\theta_2$, $\varphi$

Initialize target networks $\theta_1' \leftarrow \theta_1$, $\theta_2' \leftarrow \theta_2$, $\varphi' \leftarrow \varphi$

Initialize replay buffer $B$

Initialize state buffer $S$ and $KDTree$

**for** $t = 1$ **to** $T$ **do**

  Select action with exploration noise $a \sim \pi_\varphi(s) + \epsilon$, $\epsilon \sim N(0, \sigma)$ and observe reward $r$ and new state $s'$

  Store transition tuple $(s, a, r, s')$ in $B$

  **if** Novelty $(s', S, KDTree)$ **then**

   $r_e \leftarrow$ Compute exploration reward

   Store state $s'$ in $S$ and update $KDTree$

  **end if**

  Sample mini-batch of $N$ transitions $(s, a, r, s')$ from $B$

  $a' \leftarrow \pi_{\varphi'}(s') + \epsilon$, $\epsilon \sim clip(N(0, \sigma'), -c, c)$

  $y \leftarrow r + \gamma \min_{i=1,2} Q_{\theta_i'}(s', a')$

  Update critics $\theta_i \leftarrow argmin_{\theta_i} N^{-1} \sum (y - Q_{\theta_i}(s, a))^2$

  **if** $t$ mod $d == 0$ **then**

   Update $\varphi$ by the deterministic policy gradient:

   $\nabla_\varphi J(\varphi) = N^{-1} \nabla_a Q_{\theta_1}(s, a)|_{a=\pi_\varphi(s)} \nabla_\varphi \pi_\varphi(s)$

   Update target networks:

   $\theta_i' \leftarrow \tau \theta_i + (1 - \tau) \theta_i'$

   $\varphi' \leftarrow \tau \varphi + (1 - \tau) \varphi'$

  **end if**

**end for**

## IV. EXPERIMENTS

In this section, we evaluate the performance of the TD3 algorithm with the integrated EECL module. We describe our experimental setup, present the results, and discuss their implications.

### A. Environment

We conducted our experiments in the Robosuite Panda Lift task environment, a simulated robotic manipulation task where a 7-DOF robotic arm is trained to lift a cube from a table, as illustrated in Fig. 1.

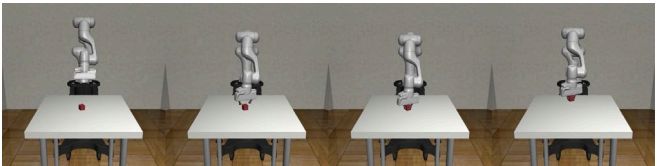

Fig. 1. Robosuite Panda Block Lifting environment.

### B. Results

Visual inspection of the agent's behavior as shown in Fig. 2. shows that the EECL-enhanced TD3 algorithm enables the robotic arm to explore a wider range of strategies for lifting the cube, leading to more robust and adaptable policies. The baseline TD3, in contrast, tends to converge to suboptimal strategies that do not generalize well to variations in the task.

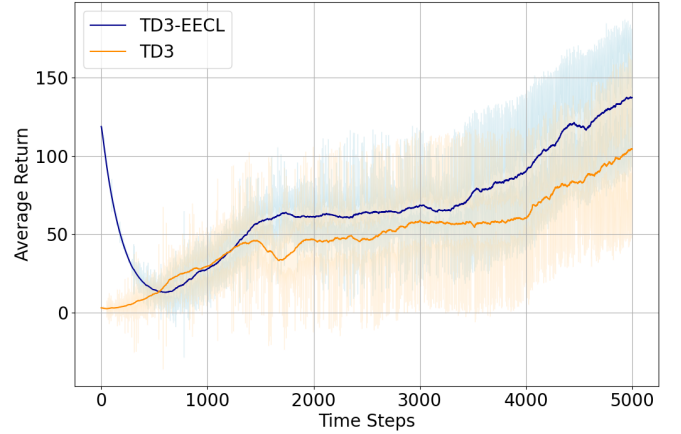

Fig. 2. Learning curves comparsion of TD3 with EECL versus TD3 for Robosuite Panda Block Lifting task. The shaded area indicates half of the standard deviation of the average evaluation across 5 trials. The curves are uniformly smoothed for better visual clarity.

The experimental results clearly demonstrate the advantages of incorporating the EECL module into the TD3 algorithm. These benefits are evident in several key performance metrics:

- **Average Cumulative Reward:** The EECL-enhanced TD3 algorithm significantly outperforms the baseline TD3 in terms of average cumulative reward. The improvement is consistently observed across different random seeds, indicating the robustness of our approach.

- **Convergence Speed:** The EECL-enhanced TD3 demonstrates a faster convergence speed compared to the baseline. The agent with the EECL module reaches stable performance levels in fewer episodes, highlighting the effectiveness of the enhanced exploration strategy.

- **Exploration Efficiency:** The EECL module significantly improves the exploration efficiency. The rate of novel state discovery is higher in the EECL-enhanced TD3, leading to more diverse experiences and better policy optimization.

### C. Evaluation

To assess our algorithm, we evaluate its performance on the robosuite Panda Lift task, utilizing the GymWrapper for interfacing. The environment settings are kept unchanged to ensure a fair comparison.

Our TD3 implementation employs a two-layer feedforward neural network with 400 and 300 hidden nodes respectively for the critic network, and 400 and 300 hidden nodes for the actor network, both using rectified linear units (ReLU) between each layer. The critic network receives both state and action as input to the first layer. We update the network parameters using the Adam optimizer with a learning rate of 0.001 for the actor and the AdamW optimizer with a learning rate of 0.001 for the critic, along with a weight decay of 0.005. After each time step, the networks are trained using a mini-batch of 128 transitions sampled uniformly from a replay buffer with a maximum size of 1,000,000 transitions.

The exploration-enhanced contrastive learning (EECL) module is integrated into the TD3 framework. The EECL module includes several parameters: state dimension (state_dim), distance threshold for novel state identification (threshold set to 0.1), initial maximum exploration reward (max_exploration_reward set to 0.75), reward decay factor (reward_decay set to 0.997), and maximum states stored (max_states set to 1000). The KDTree mechanism is utilized to manage the exploration states, where new states are added and old states are pruned to maintain the maximum state limit. Novel states are identified based on Euclidean distance compared to historical data, ensuring efficient exploration.

Target policy smoothing is achieved by adding Gaussian noise (mean = 0, standard deviation = 0.2) to the actions chosen by the target actor network, which is clipped to the range [-0.5, 0.5]. Delayed policy updates are used, where the actor and target critic networks are updated every two iterations. Target networks are updated with a soft update parameter, $\tau$, set to 0.005.

To mitigate the dependency on the initial policy parameters, we employ a purely exploratory policy for the first 1000 time steps. Following this, an off-policy exploration strategy is adopted by adding Gaussian noise with a standard deviation of 0.1 to each action.

Each task is executed for 5000 time steps. During evaluations, the average reward over 10 episodes without exploration noise is reported. Our results are averaged over 5 random seeds to ensure robustness.

We compare our algorithm against the baseline TD3 to evaluate the impact of the EECL module on exploration efficiency, convergence speed, and average cumulative reward. The results show that the EECL-enhanced TD3 surpasses the baseline in all metrics, demonstrating significant improvements in both efficiency and stability. Specifically, the EECL module effectively accelerates the learning process, allowing the agent to reach optimal performance more quickly. This enhancement not only validates the module's utility in complex control tasks but also highlights its potential for broader applications in reinforcement learning.

## V. Conclusion

In this paper, we addressed the issue of suboptimality in Twin Delayed Deep Deterministic Policy Gradient (TD3) when applied to the control of 7-DOF robotic arms. To tackle the challenge of insufficient exploration, which often leads to suboptimal policies, we proposed a novel Exploration-Enhanced Contrastive Learning (EECL) module. This module enhances exploration by providing additional rewards for encountering novel states. It effectively stores previously explored states in a buffer, identifies new states by comparing them with historical data using a KDTree framework, and assigns exploration rewards based on the novelty of the states encountered.

Our results demonstrate that integrating the EECL module into the TD3 algorithm significantly improves both performance and stability. The EECL-enhanced TD3 algorithm shows a marked increase in average cumulative reward, faster convergence speed, and greater exploration efficiency compared to the baseline TD3. These improvements are consistently observed across different random seeds, indicating the robustness and reliability of our approach. The enhanced exploration strategy not only accelerates the learning process but also leads to more diverse experiences and better policy optimization, highlighting the effectiveness of our method in overcoming the exploration challenges inherent in reinforcement learning for robotic control tasks.

Our findings suggest that the EECL module is a valuable addition to other reinforcement learning algorithms, enhancing exploration effectively. Future work will apply the EECL module in various environments, including complex tasks, to validate its versatility. Additionally, we plan to integrate EECL with other RL algorithms like PPO and SAC to evaluate its broader applicability. This research aims to refine the EECL module, contributing to more efficient reinforcement learning strategies for complex robotic control and beyond.

ACKNOWLEDGMENT

The work described herein was conducted in the Vibrations Robotics Mechatronics Lab of Department of Power Mechanical Engineering at National Tsing Hua University. Lab members' invaluable support and guidance are much appreciated.